\documentclass[runningheads]{llncs}
\usepackage[T1]{fontenc}
\usepackage{hyperref}
\hypersetup{
    colorlinks,%
    citecolor=blue,%
    filecolor=black,%
    linkcolor=blue,%
    urlcolor=blue
}

\usepackage{amsmath}
\usepackage{amssymb}
\usepackage{booktabs}
\usepackage{xcolor}
\usepackage{csquotes}
\usepackage[english]{babel} 
\usepackage{graphicx}
\begin{document}
\title{Efficient Extractive Text Summarization for Online News Articles Using Machine Learning}
%
%
\author{Sajib Biswas\inst{1}\orcidID{0009-0002-3871-1062} \and
Milon Biswas\inst{2}\orcidID{0000-0002-2825-8308} \and
Arunima Mandal\inst{1}\
\and
Fatema Tabassum Liza\inst{1}
\and
Joy Sarker\inst{3} }
\authorrunning{S. Biswas et al.}
\titlerunning{Efficient Extractive Text Summarization}
%
\institute{Florida State University, 222 S Copeland St, Tallahassee, FL, 32306, United States 
\email{\{sbiswas, am19cf, fl19g\}@fsu.edu} \and
Towson University, 8000 York Rd, Towson, MD, 21252, United States\\
\email{milonbiswas4702@gmail.com}
\and
Bangladesh Agricultural University, BAU Main Road, Mymensingh, 2202, Bangladesh\\
\email{joy.22220203@bau.edu.bd}}
\titlerunning{Efficient Extractive Summarization for Online News}
\authorrunning{S. Biswas et al.}
\maketitle              
%
\begingroup
\renewcommand\thefootnote{}%
\footnotetext{\fbox{\parbox{\dimexpr\linewidth-2\fboxsep-2\fboxrule\relax}{%
This is a post-peer-review, pre-copyedit version of a paper accepted for publication at the 5th International Conference on Applied Intelligence and Informatics 2025 (AII 2025). The final authenticated version will be available online in the conference proceedings published by Springer.
}}}
\endgroup
\begin{abstract}
In the age of information overload, content management for online news articles relies on efficient summarization to enhance accessibility and user engagement. This article addresses the challenge of extractive text summarization by employing advanced machine learning techniques to generate concise and coherent summaries while preserving the original meaning. Using the Cornell Newsroom dataset, comprising 1.3 million article-summary pairs, we developed a pipeline leveraging BERT embeddings to transform textual data into numerical representations. By framing the task as a binary classification problem, we explored various models, including logistic regression, feed-forward neural networks, and long short-term memory (LSTM) networks. Our findings demonstrate that LSTM networks, with their ability to capture sequential dependencies, outperform baseline methods like Lede-3 and simpler models in F1 score and ROUGE-1 metrics. This study underscores the potential of automated summarization in improving content management systems for online news platforms, enabling more efficient content organization and enhanced user experiences.
\end{abstract}

\section{Introduction}

Automatic text summarization refers to the process of generating a concise and coherent version of a longer document while preserving its core meaning~\cite{el2021automatic}. With the exponential growth of textual data on the internet, it has become increasingly challenging for users to efficiently consume and extract relevant information~\cite{vilca2017study}. Although manual summarization remains effective, it is time-consuming and requires substantial human effort. Consequently, the demand for automatic summarization systems has surged, driven by the over-abundance of online content such as news articles, research papers, and social media posts~\cite{wibawa2024survey}.

Text summarization has found applications across various domains, including journalism, analysis of legal documents, and financial document processing~\cite{rawte2020comparative}. For example, when browsing a news portal, users often scan headlines before deciding whether to read further. News article summarization, therefore, has become a prominent area of research. Recent advancements in deep learning have significantly improved the quality of AI generated summaries~\cite{zhang2024systematic}. As digital news platforms become more prevalent, there is a growing need for effective methods to manage and summarize news content, ensuring users receive concise and meaningful information without feeling overwhelmed.

Summarization techniques are broadly categorized into two types: extractive and abstractive. Extractive summarization methods identify and select the most relevant sentences from the original document to construct a summary. These methods often employ ranking algorithms, graph-based techniques, and machine learning models to determine sentence importance~\cite{erkan2004lexrank,mihalcea2004textrank}. In contrast, abstractive summarization generates novel sentences that express the key ideas of the document. This approach typically relies on neural network based architectures such as sequence-to-sequence models and transformers~\cite{vaswani2017attention}. While abstractive methods can produce fluent and human-like summaries, they face challenges including factual inconsistencies and the need for large-scale training data~\cite{kryscinski2019evaluating}. Given these difficulties, extractive summarization remains a robust and widely adopted strategy, especially for structured texts like news articles~\cite{harinatha2021evaluating}.

In this work, we apply multiple techniques to perform extractive summarization of online news content and assess the quality of the generated summaries using standard evaluation metrics. The main contributions of this paper are as follows:
\begin{itemize}
    \item We use the Cornell Newsroom dataset~\cite{grusky2018newsroom}, which comprises $1.3$ million real-world news articles and their corresponding summaries, sourced from $39$ major publications.
    \item We implement and train a number of machine learning models
    on the dataset to generate extractive summaries and assess their performance.
    \item We empirically demonstrate that long short-term memory based models outperform traditional baselines such as Lede-3 and other machine learning models in terms of metrics such as F1 and ROUGE-1 score.
\end{itemize}

\section{Background}

Extractive text summarization has advanced significantly in recent years, particularly with the integration of machine learning techniques. In this section, we highlight a few notable works that have contributed to the development of extractive summarization methods.

Moratanch and Chitrakala~\cite{moratanch2017survey} provided a comprehensive overview of extractive summarization models, categorizing them into supervised and unsupervised approaches. Among unsupervised methods, Neto et al.~\cite{neto2002automatic} proposed a technique that learns classification probabilities to distinguish summary sentences from non-summary ones.  Hingu et al.~\cite{hingu2015automatic} introduced a neural network based model that integrates sentence features, ranks sentences, and selects the most salient ones for summarization. Since unsupervised approaches do not require manually created summaries, they are particularly effective for handling large-scale datasets.

Suanmali et al.~\cite{suanmali2009fuzzy} made a notable contribution in the field of supervised methods, who utilized fuzzy systems to preprocess text features before extracting key sentences. However, their method faces challenge in handling dangling references. Further improvements were proposed by Sankarasubramaniam
et al.~\cite{sankarasubramaniam2014text}, who combined graph-based and concept-based techniques by constructing a binary classification graph using Wikipedia concepts and applying an iterative ranking algorithm to extract the most relevant sentences.

The evolution of automatic text summarization has progressed from traditional statistical approaches to advanced machine learning models~\cite{widyassari2022review}. Recent surveys highlight the increasing use of machine learning models to enhance summarization performance~\cite{zhang2024systematic}. Dheer and Dhankhar~\cite{dheer2023automatic} conducted a comparative study of extractive and abstractive summarization techniques for news articles, evaluating various deep learning models on the CNN/ DailyMail dataset. Their study assessed summary quality using ROUGE scores and offered valuable insights into the relative effectiveness of different methods. However, it has been shown that deep learning-based models are vulnerable to adversarial perturbations, which may influence their summarization outputs~\cite{biswas2025adversarial,chacko2024adversarial}.


Beyond general applications, extractive summarization has also proven to be valuable in a few specific domains. For example, Bui et al.~\cite{bui2016extractive} developed an extractive summarization technique to assist in data extraction from full-text biomedical articles. 
A recent study leverages BERT~\cite{devlin2018bert} embeddings, combined with K-Means clustering to identify sentences closest to the centroid for summary generation~\cite{miller2019leveraging} and demonstrates the effectiveness of generated summaries for educational purpose.

\section{Methodology}

\begin{figure}[ht]
    \centering
    \includegraphics[width=0.50\columnwidth]{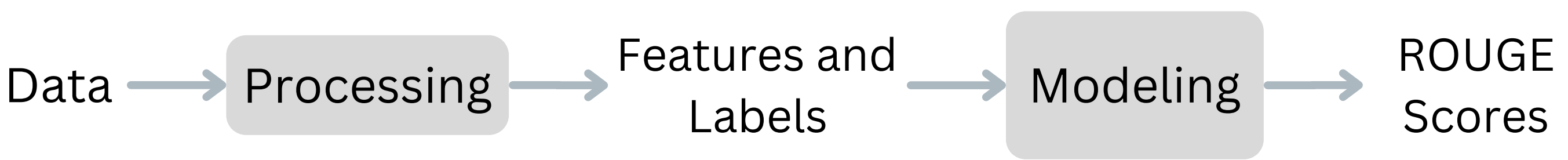}
    \caption{Overall pipeline}
    \label{fig:overall_pipeline}
\end{figure}

In this section, we describe our experimental pipeline and the machine learning models used for extractive summarization, organized step by step in the following subsections. A high-level overview of our approach is presented in Figure~\ref{fig:overall_pipeline}. We begin by processing the dataset to generate features and labels, which are then used to train supervised machine learning models. We construct a machine learning pipeline in which sentence embeddings are generated using BERT~\cite{devlin2018bert}, a pretrained language model based on transformer. The performance of these models are evaluated using ROUGE scores, a standard metric that measures the overlap of words between the predicted and human-written gold summaries.

\subsection{Data Processing Pipeline}

\begin{figure*}[ht]
   \centering
    \includegraphics[width=\textwidth]{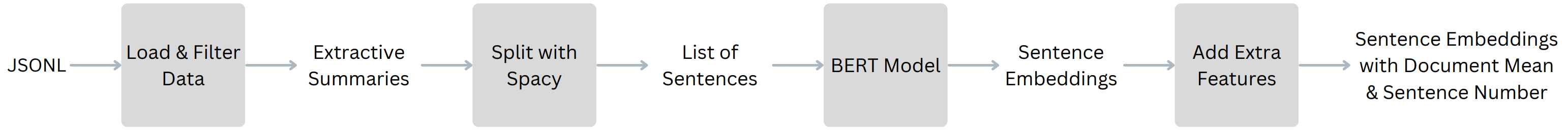}
    \caption{Data preparation and processing pipeline.}
    \label{fig:preprocessing_pipeline}
\end{figure*}

The first step involves transforming the raw textual data into a format suitable for machine learning models. We load and filter the Newsroom dataset~\cite{grusky2018newsroom}, which is provided in JSONL format, to retain only samples with extractive summaries. Each article-summary pair is segmented into sentences using \textit{SpaCy}, and sentence-level embeddings are generated using the BERT model. BERT generates embeddings for each sentence, which are represented as dense, multidimensional vectors with interesting geometric properties~\cite{biswas2022geometric}. Additional features such as sentence position and document-level context are concatenated to enrich the representation space. These attributes help the model understand both the identity of the document and the relative position of each sentence within it. The data preparation and processing pipeline is illustrated in Figure~\ref{fig:preprocessing_pipeline}.

BERT, which stands for Bidirectional Encoder Representations from Transformers, is a language model pretrained on large-scale text corpora. It can be fine-tuned for various downstream tasks. For instance, BERT embeddings can be fed into a classifier to detect whether a given sentence belongs to a summary. We frame the extractive summarization task as a binary classification problem, where the goal is to predict whether each sentence in an article should be included in the summary.

After generating sentence-level embeddings and assigning binary labels, the data is ready to be fed into supervised machine learning models for summary extraction.

\subsection{Machine Learning Pipeline}

\begin{figure}[htbp]
    \centering
    \includegraphics[width=0.70\columnwidth]{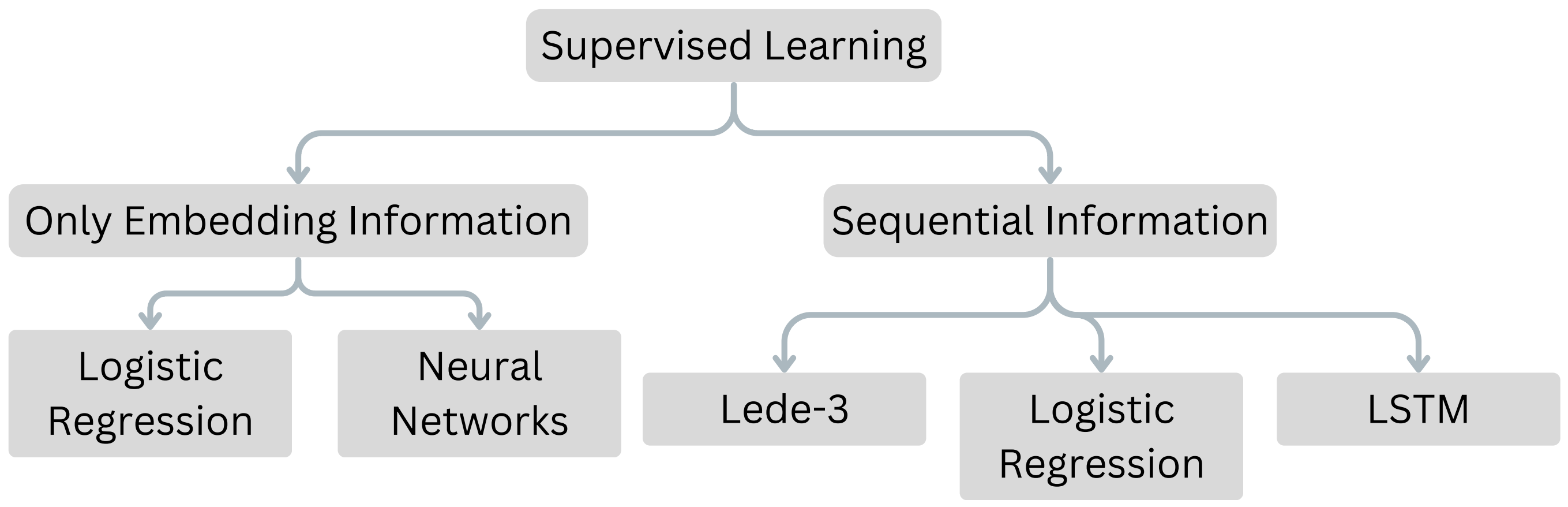}
    \caption{Machine learning models we use for extractive summarization.}
    \label{fig:supervised_structure}
\end{figure}

This section outlines the machine learning models used in our experiments, along with the motivation behind their selection. The different models we use are illustrated in Figure~\ref{fig:supervised_structure}.

We approach the problem using two strategies: \textbf{(1)} models that rely solely on sentence-level embeddings, and 
\textbf{(2)} models that incorporate sequential information of a sentence with its embeddings.

In the first approach, we use logistic regression and fully-connected neural networks. Logistic regression is a widely-used baseline for binary classification problems. We employ the logistic regression implementation from Python’s \textit{Keras} framework. For neural networks, we experiment with multiple architectures and hyperparameters, training each model for 50 epochs. Our best-performing architecture includes three dense layers, which yielded the most optimal results based on the evaluation metrics.

For the second approach, which leverages sequential information, we evaluate the baseline method known as Lede-3 with models such as logistic regression (with positional features), and long short-term memory (LSTM) networks. LSTM is a variant of Recurrent Neural Networks (RNNs) designed to capture long-range dependencies in sequential data. We experiment with both uni-directional (LSTM Uni) and bi-directional (LSTM Bi) variants. LSTMs are well-suited for textual data because they can preserve contextual information by passing hidden states across time steps. In the following subsections, we describe the baseline method and the machine learning models in more detail.

\subsubsection{Lede-3} 
A common automatic summarization strategy of online publications is to copy the first few sentences of the text and treat that as a summary. Lede-3 is a widely used baseline in automatic news summarization, in which the first three sentences of the text are selected as the summary~\cite{see-etal-2017-get}. Despite its simplicity, Lede-3 performs very well and is competitive with more sophisticated summarization techniques~\cite{nallapati2016classify}. This approach is remarkably effective for news datasets like CNN/Daily Mail~\cite{hermann2015teaching} and the Newsroom corpus, as journalistic writing often presents the most critical information at the beginning of the article~\cite{grusky2018newsroom}. Because of its strong performance, Lede-3 remains a standard baseline for evaluating new summarization models.

\subsubsection{Logistic Regression} 
Logistic regression is a well-established statistical method for binary classification tasks~\cite{kleinbaum2002logistic}. It serves as a strong baseline for extractive summarization by determining whether each sentence in a document should be included in the summary~\cite{peng2002introduction}. 
Let \( \mathbf{x} \in \mathbb{R}^n \) represent the feature vector of a sentence (e.g., BERT embedding, positional encoding), and \( y \in \{0, 1\} \) represent the binary label ($0$ for non-summary and $1$ for summary sentence). The logistic regression model estimates the probability \( P(y=1 \mid \mathbf{x}) \) using the sigmoid function:

\[
P(y = 1 \mid \mathbf{x}) = \sigma(\mathbf{w}^\top \mathbf{x} + b) = \frac{1}{1 + \exp(-(\mathbf{w}^\top \mathbf{x} + b))}
\]
where:
\begin{itemize}
  \item \( \mathbf{w} \in \mathbb{R}^n \) is the weight vector,
  \item \( b \in \mathbb{R} \) is the bias term,
  \item \( \sigma(\cdot) \) is the sigmoid (logistic) function.
\end{itemize}
The model is trained by minimizing the binary cross-entropy loss function over a training set of \( m \) examples:
\[
\mathcal{L}(\mathbf{w}, b) = -\frac{1}{m} \sum_{i=1}^{m} \left[ y^{(i)} \log(\hat{y}^{(i)}) + (1 - y^{(i)}) \log(1 - \hat{y}^{(i)}) \right]
\]
where:
\begin{itemize}
  \item \( y^{(i)} \) is the true label for the \(i\)-th example,
  \item \( \hat{y}^{(i)} = \sigma(\mathbf{w}^\top \mathbf{x}^{(i)} + b) \) is the predicted probability.
\end{itemize}

The optimization is typically performed using \textit{gradient descent} or its adaptive variants, such as Adam. The model’s decision boundary is linear, which means it can be effective when the summary-worthy sentences are linearly separable in the feature space. Despite its simplicity, logistic regression shows competitive performance in various summarization benchmarks, particularly when integrated with high-quality semantic information, such as sentence embeddings, positional encodings, TF-IDF scores, and similarity with the title or lead sentence~\cite{li2023logistic}. Its interpretability and effectiveness make it a strong baseline in extractive summarization research.

\subsubsection{Feed-forward Neural Network}
Feed-forward neural networks (NNs), also known as dense neural networks, are composed of multiple fully connected layers where information flows in a single direction, from input to output~\cite{rosenblatt1958perceptron}. In extractive summarization, NNs can be used to classify whether a sentence belongs to the summary, given its embedding and additional features such as position and importance score. Unlike logistic regression, these models can learn non-linear decision boundaries, making them more expressive. For example, Zhang et al.~\cite{zhang2018neural} showed that NNs can effectively classify sentence-level importance in documents, achieving improved summarization results over linear models.

A feed-forward network consists of multiple layers of neurons where each neuron in a layer is connected to every neuron in the previous layer. There is no cycle or feedback loop in this architecture, which distinguishes it from recurrent neural networks~\cite{rumelhart1986learning}. Information flows from the input layer, through one or more hidden layers, to the output layer.

Let \( \mathbf{x} \in \mathbb{R}^n \) represent the input feature vector for a sentence. The computation in the first hidden layer can be written as:
\[
\mathbf{h}^{(1)} = f^{(1)}(\mathbf{W}^{(1)} \mathbf{x} + \mathbf{b}^{(1)})
\]
where,
\begin{itemize}
  \item \( \mathbf{W}^{(1)} \in \mathbb{R}^{d_1 \times n} \) is the weight matrix for the first layer,
  \item \( \mathbf{b}^{(1)} \in \mathbb{R}^{d_1} \) is the bias vector,
  \item \( f^{(1)} \) is a non-linear activation function, such as ReLU or tanh,
  \item \( \mathbf{h}^{(1)} \in \mathbb{R}^{d_1} \) is the output of the first hidden layer.
\end{itemize}
Subsequent layers follow a similar pattern. For a network with \( L \) layers, the general transformation at layer \( l \) is given by:
\[
\mathbf{h}^{(l)} = f^{(l)}(\mathbf{W}^{(l)} \mathbf{h}^{(l-1)} + \mathbf{b}^{(l)})
\]
The final layer is a single neuron with a sigmoid activation function to deduce the probability whether a sentence belongs to the summary:
\[
\hat{y} = \sigma(\mathbf{w}^{(L)} \cdot \mathbf{h}^{(L-1)} + b^{(L)})
\]
where \( \hat{y} \in [0, 1] \) is the predicted probability, and \( \sigma(z) = \frac{1}{1 + e^{-z}} \) is the sigmoid function.
The model is trained using the binary cross-entropy loss:
\[
\mathcal{L} = -\frac{1}{m} \sum_{i=1}^{m} \left[ y^{(i)} \log(\hat{y}^{(i)}) + (1 - y^{(i)}) \log(1 - \hat{y}^{(i)}) \right]
\]
where:
\begin{itemize}
  \item \( m \) is the number of training examples,
  \item \( y^{(i)} \in \{0, 1\} \) is the true label for the \(i\)-th sentence,
  \item \( \hat{y}^{(i)} \) is the predicted probability.
\end{itemize}

Feed-forward neural networks are more expressive than linear models like logistic regression and can capture complex relationships among features. However, they do not capture sequential dependencies between sentences, which may limit their effectiveness when context is important. Nonetheless, when combined with strong semantic features such as contextual embeddings from BERT, neural networks serve as a competitive baseline for extractive summarization.

\subsubsection{Long Short-Term Memory (LSTM)}
Recurrent neural networks (RNNs) are designed for sequential data processing, maintaining a hidden state that captures information from previous time steps.~\cite{rumelhart1986learning}. This makes them well-suited for sequential tasks such as language modeling, time series prediction, and text summarization. For a sequence of inputs \( \mathbf{x}_1, \mathbf{x}_2, \ldots, \mathbf{x}_T \), the hidden state \( \mathbf{h}_t \) at time step \( t \) in a standard RNN is updated as:
\[
\mathbf{h}_t = \tanh(\mathbf{W}_h \mathbf{h}_{t-1} + \mathbf{W}_x \mathbf{x}_t + \mathbf{b})
\]
where:
\begin{itemize}
  \item \( \mathbf{h}_{t-1} \in \mathbb{R}^d \) is the hidden state from the previous time step,
  \item \( \mathbf{x}_t \in \mathbb{R}^n \) is the input at time \( t \),
  \item \( \mathbf{W}_h \), \( \mathbf{W}_x \) are weight matrices,
  \item \( \mathbf{b} \) is the bias vector,
  \item \( \tanh \) is the hyperbolic tangent activation function.
\end{itemize}

Although RNNs can model temporal dependencies, they suffer from vanishing gradient problem, which limits their ability to capture long-range dependencies in sequences. This limitation severely restricts their effectiveness in tasks that require long-range contextual information such as processing large documents. To overcome these challenges, long short-term memory (LSTM) networks were introduced by Hochreiter and Schmidhuber~\cite{hochreiter1997long}. LSTMs introduce memory cells and gating mechanisms that regulate the flow of information, allowing them to retain relevant information over long sequences. An LSTM cell consists of the following components:
{\small
\begin{align*}
\mathbf{f}_t &= \sigma\left(\mathbf{W}_f \mathbf{x}_t + \mathbf{U}_f \mathbf{h}_{t-1} + \mathbf{b}_f\right) &\text{(forget gate)} \\
\mathbf{i}_t &= \sigma\left(\mathbf{W}_i \mathbf{x}_t + \mathbf{U}_i \mathbf{h}_{t-1} + \mathbf{b}_i\right) &\text{(input gate)} \\
\tilde{\mathbf{c}}_t &= \tanh\left(\mathbf{W}_c \mathbf{x}_t + \mathbf{U}_c \mathbf{h}_{t-1} + \mathbf{b}_c\right) &\text{(candidate cell state)} \\
\mathbf{c}_t &= \mathbf{f}_t \odot \mathbf{c}_{t-1} + \mathbf{i}_t \odot \tilde{\mathbf{c}}_t &\text{(updated cell state)} \\
\mathbf{o}_t &= \sigma\left(\mathbf{W}_o \mathbf{x}_t + \mathbf{U}_o \mathbf{h}_{t-1} + \mathbf{b}_o\right) &\text{(output gate)} \\
\mathbf{h}_t &= \mathbf{o}_t \odot \tanh\left(\mathbf{c}_t\right) &\text{(updated hidden state)}
\end{align*}
}
where:
\begin{itemize}
  \item \( \sigma \) is the sigmoid activation function,
  \item \( \odot \) denotes element-wise multiplication,
  \item \( \mathbf{x}_t \) is the input vector at time \( t \),
  \item \( \mathbf{f}_t,\mathbf{i}_t,\mathbf{o}_t \) are gate activations at time \( t \),
  \item \( \mathbf{W}_{\{\cdot\}} \) are the input-to-gate weight matrices,
  \item \( \mathbf{U}_{\{\cdot\}} \) are the hidden-to-gate weight matrices,
  \item \( \mathbf{b}_{\{\cdot\}} \) are the bias vectors for each gate.
\end{itemize}

The gating mechanisms enable LSTMs to learn what information to retain, update, or discard, allowing them to capture dependencies over long sequences and overcome the limitations of standard RNNs.

In extractive text summarization, LSTMs are used to model the sequential relationships between sentences in a document. 
Each sentence is represented by an embedding generated by a language model (e.g., BERT or GloVe) and is then fed into the LSTM network. The network outputs a probability for each sentence, indicating its likelihood of being included in the summary. By leveraging contextual dependencies between sentences, LSTMs are able to produce more coherent and context-aware summaries than feed-forward architectures. 

Cheng and Lapata~\cite{cheng2016neural} introduced one of the first end-to-end LSTM-based extractive summarization models, where sentences are encoded with a bidirectional LSTM and scored jointly using sequence prediction. This was extended by  Nallapati et al.~\cite{nallapati2017summarunner} by the incorporating sentence-level features and reinforcement learning. More recent works have explored combining LSTMs with attention mechanisms and contextual embeddings to further improve performance of text summarization~\cite{zhong2020extractive}.

\subsection{Evaluation Metrics}

In this section, we discuss the metrics we use to evaluate different text summarization methods. To evaluate the results, we use both classification-oriented metrics (Precision, Recall, and F1-score) and content-overlap metrics (ROUGE-1).

Since the sentences in the reference (gold) summaries are not always exact matches with those in the source article, we compute the similarity between each document sentence and each summary sentence using cosine similarity~\cite{singhal2001modern}. Based on this metric, sentences that sufficiently match summary sentences are assigned a label of \textbf{1}, while others are labeled as \textbf{0}. To quantify classification performance, we use the confusion matrix, which provides a tabular representation of predicted versus actual labels. For binary classification, the structure is shown in Table~\ref{tab:confusion_matrix}, where each row of the confusion matrix represents the actual class, and each column represents the predicted class. 

\begin{table}[htbp]
\centering
\small
\caption{Confusion matrix for binary classification}
\begin{tabular}{lcc}
\toprule
\textbf{Actual / Predicted} & \textbf{Positive (1)} & \textbf{Negative (0)} \\
\midrule
\textbf{Positive (1)} & True Positive (TP) & False Negative (FN) \\
\textbf{Negative (0)} & False Positive (FP) & True Negative (TN) \\
\bottomrule
\end{tabular}
\label{tab:confusion_matrix}
\end{table}

From the confusion matrix, we compute the following standard classification metrics:

\[
\text{Recall} = \frac{\text{TP}}{\text{TP} + \text{FN}}
\]
Recall (Sensitivity or True Positive Rate) measures the ratio of actual positive instances that the model correctly identifies.

\[
\text{Precision} = \frac{\text{TP}}{\text{TP} + \text{FP}}
\]
Precision measures the ratio of predicted positive instances that are actually positive.
Since precision and recall are often in tension and improving one reduces the other, we also compute the F1-score, their harmonic mean:
\[
\text{F1-Score} = 2 \cdot \frac{\text{Precision} \cdot \text{Recall}}{\text{Precision} + \text{Recall}}
\]

For text summarization, we also compute the ROUGE-N metrics, which have become the commonly used standard for measuring content overlap between system-generated and human-written summaries~\cite{lin2004rouge}. ROUGE (Recall-Oriented Understudy for Gisting Evaluation) compares the \textit{n}-gram overlaps between the system-generated summary and one or more human-written reference summaries~\cite{graham2015re}. The ROUGE-N recall score is formally defined as:
\[
\text{ROUGE-N} = \frac{\sum_{S \in \text{RefSumm}} \sum_{\text{gram}_n \in S} \text{Count}_{\text{match}}(\text{gram}_n)}{\sum_{S \in \text{RefSumm}} \sum_{\text{gram}_n \in S} \text{Count}(\text{gram}_n)}
\]
where:
\begin{itemize}
    \item \( \text{gram}_n \) denotes any contiguous sequence of \( n \) words (an \textit{n}-gram),
    \item \( \text{Count}_{\text{match}}(\text{gram}_n) \) is the number of matching \( n \)-grams in both the system and reference summaries,
    \item \( \text{Count}(\text{gram}_n) \) is the total number of \( n \)-grams in the reference summaries.
\end{itemize}
Although ROUGE-N primarily measures recall, precision and F1-score variants can also be computed. ROUGE-1 and ROUGE-2, which measure unigram and bigram overlaps respectively, are among the most commonly used in summarization research.

\section{Experimental Results}

In this section, we present the results of our experiments on extractive summarization using various supervised learning models. The dataset consists of $5,000$ news articles, from which we extracted approximately $150,000$ sentences for model training and evaluation.

\subsection{Supervised Learning Using Only Embedding Information}

We first evaluate models that utilize only sentence-level embedding features, without modeling sequential information. Table~\ref{tab:nn_rouge1} reports the performance of different feed-forward (dense) neural networks with varying architectures. The two-layer network with 50 neurons per layer (NN 50 50 Bal) achieves the highest F1-score of $0.415$.

\begin{table}[htbp]
\centering
\small
\caption{ROUGE-1 scores for feed-forward neural network models}
\begin{tabular}{lccc}
\toprule
\textbf{Model} & \textbf{F1} & \textbf{Recall} & \textbf{Precision} \\
\midrule
NN 25 25 Bal & 0.403 & 0.379 & 0.577 \\
NN 25 50 Bal & 0.396 & 0.390 & 0.551 \\
NN 50 50 Bal & \textbf{0.415} & 0.397 & 0.583 \\
\bottomrule
\end{tabular}
\label{tab:nn_rouge1}
\end{table}

We also compare the best-performing neural network model (NN 50 50 Bal) to a logistic regression classifier trained on the same embedding features. As shown in Table~\ref{tab:rouge1_logreg_nn}, logistic regression slightly outperforms the neural network in terms of F1-score, achieving a score of $0.416$. This demonstrates that simple linear models can still perform competitively when provided with high-quality sentence embeddings.

\begin{table}[htbp]
\centering
\small
\caption{ROUGE-1 scores for logistic regression and neural network}
\begin{tabular}{lccc}
\toprule
\textbf{Model} & \textbf{F1} & \textbf{Recall} & \textbf{Precision} \\
\midrule
Logistic Reg & \textbf{0.416} & 0.398 & 0.578 \\
NN 50 50 Bal & 0.415 & 0.397 & 0.583 \\
\bottomrule
\end{tabular}
\label{tab:rouge1_logreg_nn}
\end{table}

\subsection{Supervised Learning Using Sequential Information}

To capture contextual and sequential relationships between sentences, we evaluate models based on LSTM networks. Table~\ref{tab:lstm_rouge1} presents ROUGE-1 scores for uni-directional and bi-directional LSTM models with varying hidden layer sizes. The BiLSTM model with $50$ and $75$ neurons achieves the highest F1-score of $0.599$, demonstrating the advantage of capturing the context in both directions for sentence classification tasks.

\begin{table}[htbp]
\centering
\small
\caption{ROUGE-1 Scores for LSTM Models}
\begin{tabular}{lccc}
\toprule
\textbf{Model} & \textbf{F1} & \textbf{Recall} & \textbf{Precision} \\
\midrule
LSTM Uni 25 & 0.596 & 0.579 & 0.778 \\
LSTM Uni 50 & 0.596 & 0.576 & 0.783 \\
LSTM Bi 25 & 0.596 & 0.575 & 0.781 \\
LSTM Bi 50 & \textbf{0.599} & 0.577 & 0.785 \\
LSTM Bi 75 & \textbf{0.599} & 0.577 & 0.785 \\
\bottomrule
\end{tabular}
\label{tab:lstm_rouge1}
\end{table}

We then compare the best-performing BiLSTM (Bi 50) to two baselines: Lede-3 and a default logistic regression model. As shown in Table~\ref{tab:rouge1_baseline_lstm}, the LSTM Bi 50 model outperforms both baselines, achieving higher F1 and precision scores. Although it selects only the first three sentences of a document as the summary, the Lead-3 baseline remains strong and is only slightly outperformed by the LSTM model.

\begin{table}[htbp]
\centering
\small
\caption{ROUGE-1 Scores for Baselines and LSTM Model}
\begin{tabular}{lccc}
\toprule
\textbf{Model} & \textbf{F1} & \textbf{Recall} & \textbf{Precision} \\
\midrule
LEDE3 & 0.589 & \textbf{0.588} & 0.757 \\
Logistic Reg Def & 0.525 & 0.518 & 0.696 \\
LSTM Bi 50 & \textbf{0.599} & 0.577 & \textbf{0.785} \\
\bottomrule
\end{tabular}
\label{tab:rouge1_baseline_lstm}
\end{table}

Table~\ref{tab:model_summaries} presents example summaries generated by various models alongside the original news article. The Logistic Regression summary is factual and concise, emphasizing key details such as the exorcism, Mother Teresa’s death, and her sainthood, but it omits subjective or personal expressions found in the reference summary. This results in a more detached tone, which may reduce reader engagement. The neural network output retains emotional and narrative elements of the original text, including phrases like \enquote{slept like a baby} and introduces external references (e.g., Michael Cuneo). It may improve readability but slightly deviates from the core of the article. 
The LSTM and Lead-3 outputs show a strong preference for early sentences in the article, which reflects a reliance on positional information. This phenomenon is effective for news data due to the fact that critical information often appears in the beginning, but also limited in its ability to incorporate important details that occur later in the article. While similar in structure to Lede-3, LSTM output often reorders or omits less essential phrases, which may explain its slightly higher ROUGE-1 scores. 

\begin{table*}[htbp]
\centering
\tiny
\caption{Example summaries produced by different baseline methods for the same news article}
\begin{tabular}{p{0.15\textwidth}p{0.8\textwidth}}
\toprule
\textbf{Version} & \textbf{Text} \\
\midrule
\textbf{Original} & By CORKY SIEMASZKO DAILY NEWS STAFF WRITER Mother Teresa believed she was possessed by the Devil, the archbishop of Calcutta said yesterday. So the revered nun, whom the Vatican hopes to make a saint, underwent an exorcism and afterward "slept like a baby," he said. Archbishop Henry D'Souza's bizarre revelation came as millions yesterday marked the fourth anniversary of Mother Teresa's death. But D'Souza told CNN and The Associated Press in India he truly believed that. \\
\hline
\textbf{Logistic Reg} & Archbishop Henry D'Souza's bizarre revelation came as millions yesterday marked the fourth anniversary of Mother Teresa's death. The Catholic cleric said he diagnosed the demon in Mother Teresa shortly before she had a fatal heart attack Sept. 5, 1997, and died at age 87. Mother Teresa won a Nobel Prize for her life's work, and Pope John Paul has begun the process of declaring her a saint. \\
\hline
\textbf{Neural Net} & So the revered nun, whom the Vatican hopes to make a saint, underwent an exorcism and afterward "slept like a baby," he said. The Catholic cleric said he diagnosed the demon in Mother Teresa shortly before she had a fatal heart attack Sept. 5, 1997, and died at age 87. So did Michael Cuneo, author of "American Exorcism," which hits bookstores next week. \\
\hline
\textbf{LSTM} & BY CORKY SIEMASZKO DAILY NEWS STAFF WRITER With Emily Gest Thursday, September 6th 2001, 2:23AM Mother Teresa believed she was possessed by the Devil, the archbishop of Calcutta said yesterday. So the revered nun, whom the Vatican hopes to make a saint, underwent an exorcism and afterward "slept like a baby," he said. \\
\hline
\textbf{LEDE3} & BY CORKY SIEMASZKO DAILY NEWS STAFF WRITER With Emily Gest Thursday, September 6th 2001, 2:23AM Mother Teresa believed she was possessed by the Devil, the archbishop of Calcutta said yesterday. So the revered nun, whom the Vatican hopes to make a saint, underwent an exorcism and afterward "slept like a baby," he said. \\
\bottomrule
\end{tabular}
\label{tab:model_summaries}
\end{table*}


\section{Conclusion and Future Work}
In this study, we examine extractive summarization of online news articles using several machine learning models, including logistic regression, feed-forward neural networks, and LSTM-based models. Our results demonstrate that incorporating sequential sentence information significantly improves summarization performance, with LSTM models outperforming strong baselines such as Lede-3. In future, we aim to extend this research in multiple directions. First, we want to investigate the effectiveness of LLMs for both extractive and abstractive summarization, by conducting direct comparisons between them. We also plan to extend our evaluations across diverse datasets and domains to improve generalization and robustness.
%
%
%
\bibliographystyle{splncs04}
\bibliography{references}
%
\end{document}